%% file: main.tex
\documentclass[10pt,twocolumn,letterpaper]{article}

\usepackage{iccv}              

\input{preamble}

\definecolor{iccvblue}{rgb}{0.21,0.49,0.74}
\usepackage[pagebackref,breaklinks,colorlinks,allcolors=iccvblue]{hyperref}
\usepackage{caption}

\def\paperID{11892} 
\def\confName{ICCV}
\def\confYear{2025}

\title{DIMM: Decoupled Multi-hierarchy Kalman Filter for 3D Object Tracking}

\author{Jirong Zha\\
Shenzhen International Graudate School\\
Tsinghua University\\
{\tt\small zhajirong23@mails.tsinghua.edu.cn}
\and
Yuxuan Fan\\
The Hong Kong University of \\ Science and Technology (Guang Zhou)\\
{\tt\small yfan546@connect.hkust-gz.edu.cn}
\and
Kai Li\\
Shenzhen International Graudate School\\
Tsinghua University\\
{\tt\small likai24@mails.tsinghua.edu.cn}
\and
Han Li\\
Shenzhen International Graudate School\\
Tsinghua University\\
{\tt\small h-li23@mails.tsinghua.edu.cn}
\and
Chen Gao $^{\dag}$\\
Department of Electronic Engineering\\
Tsinghua University\\
{\tt\small chgao96@gmail.com}
\and
Xinlei Chen $^{\dag}$\\
Shenzhen International Graudate School\\
Tsinghua University\\
{\tt\small chen.xinlei@sz.tsinghua.edu.cn}
\and
Yong Li\\
Department of Electronic Engineering\\
Tsinghua University\\
{\tt\small liyong07@tsinghua.edu.cn}
}

\begin{document}
\maketitle
\input{sec/0_abstract}    
\input{sec/1_intro}

\input{sec/2_related_work}

\input{sec/3_problem_formulation}
\input{sec/4_methodology}
\input{sec/5_experiments}
\input{sec/6_conclusion}
{
    \small
    \bibliographystyle{ieeenat_fullname}
    \bibliography{main}
}
\end{document}

%% file: preamble.tex
\usepackage[ruled,linesnumbered]{algorithm2e}
\usepackage{tablefootnote}
\usepackage{multirow}
\usepackage{graphicx}
\usepackage{amsthm}
\usepackage{float}
\newtheorem{proposition}{Proposition}

%% file: sec/0_abstract.tex
\begin{abstract}
State estimation is challenging for 3D object tracking with high maneuverability, as the target's state transition function changes rapidly, irregularly, and is unknown to the estimator. 
Existing work based on interacting multiple model (IMM) achieves more accurate estimation than single-filter approaches through model combination,
aligning appropriate models for different motion modes of the target object over time. 
However, two limitations of conventional IMM remain unsolved. First, the solution space of the model combination is constrained as the target's diverse kinematic properties in different directions are ignored. Second, the model combination weights calculated by the observation likelihood are not accurate enough due to the measurement uncertainty. 
In this paper, we propose a novel framework, DIMM, to effectively combine estimates from different motion models in each direction, thus increasing the 3D object tracking accuracy. First, DIMM extends the model combination solution space of conventional IMM from a hyperplane to a hypercube by designing a 3D-decoupled multi-hierarchy filter bank, which describes the target's motion with various-order linear models. Second, DIMM generates more reliable combination weight matrices through a differentiable adaptive fusion network for importance allocation rather than solely relying on the observation likelihood; it contains an attention-based twin delayed deep deterministic policy gradient (TD3) method with a hierarchical reward.
Experiments demonstrate that DIMM significantly improves the tracking accuracy of existing state estimation methods by $31.61\%\sim99.23\%$.
\end{abstract}

%% file: sec/1_intro.tex
\section{Introduction}
\label{sec:intro}

\begin{figure}[t]
  \centering
  \includegraphics[width=0.95\linewidth]{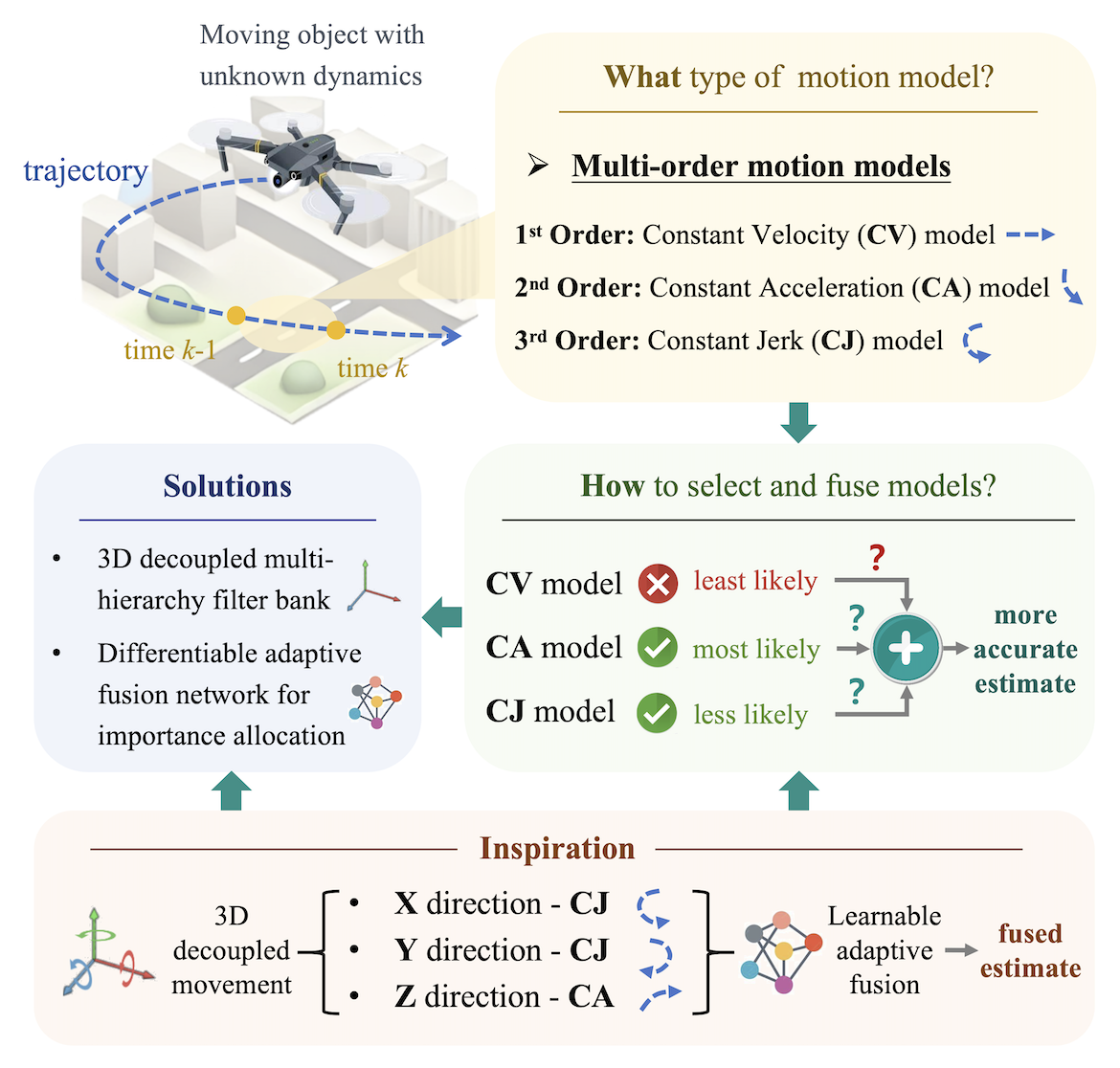}
   \caption{\textbf{Illustration of 3D object tracking with unknown dynamics.} 
   We aim to improve the estimation accuracy by determining \textit{what type of motion model} to employ and \textit{how to select and fuse the models independently of each dimension}. 
   }
   \label{fig:application}
   \vspace{-0.55cm}
\end{figure}

As a fundamental problem of perception and robotics, 3D object tracking plays a critical role in a wide range of applications such as autonomous driving \cite{li2022time3d, luo2021exploring}, urban surveillance \cite{wan2022edge}, robotic manipulation \cite{deng2020self}, target capture \cite{zha2024diffusion, zha2023privacy}, and so on \cite{yin2021center, qi2020p2b}. However, in cases where the dynamic target is highly maneuverable, the state estimation issue becomes challenging due to the unknown switching of motion models and irregular system process noises \cite{moose1979modeling}. Therefore, 
it remains less explored to tackle accurate state estimation for 3D object tracking with unknown dynamics.

Existing model-based works \cite{jiang2017multiple} commonly utilize the Interacting Multiple Model (IMM) \cite{mazor1998interacting} to deal with an object's motion uncertainties by combining various motion models in a certain ratio.
However, two major limitations of traditional IMM-based methods remain unsolved: 
\begin{itemize}[leftmargin=*]
\item Planar solution space constraint (\textbf{L1}). The traditional IMM algorithm uses direct weighting on the filters' 3D state estimate vectors, limiting the solution space of model combination as the object’s kinematic properties may vary in different directions. 
\item Observation-dependent weight instability (\textbf{L2}). The importance weights for model combination computed by observation likelihood are sensitive to measurement data's quality, since the weight values may be invalid with non-Gaussian distributed noises.
\end{itemize}

To address these two limitations, we propose a novel framework named Decoupled IMM (DIMM), to deal with accurate object tracking with unknown dynamics by expanding the combination solution space and generating adaptive combination weights. Compared to IMM, DIMM can better approximate the optimal estimate value with a more reasonable basis, \textit{i.e.}, estimate variables from different filtering models and corresponding coefficients, \textit{i.e.}, model combination weights, thus increasing the tracking accuracy.

Specifically, to overcome \textbf{L1}, we design a \textit{decoupled multi-hierarchy filter bank} composed of motion models with various orders to realize the 3D decoupling of the state estimate vector, which is theoretically proven to expand the combination solution space and facilitates subsequent independent combination of the state variable in each direction. 
To overcome \textbf{L2}, we propose a \textit{differentiable adaptive fusion network} with reinforcement learning for importance allocation of model fusion by learning the weight matrix from data. Specifically, we improve motion pattern recognition accuracy by independently weighting the estimated variable in each direction. 



%
Our contributions can be summarized as follows.
\begin{itemize}
    \item We propose a novel 3D object tracking framework, Decoupled IMM (DIMM), to improve the state estimation accuracy of dynamic objects with high maneuverability by adaptive learning-based fusion of state variables of each dimension independently. 
    \item We design a 3D decoupled multi-hierarchy filter bank to realize the independent linear combination of models' states in different dimensions. We further propose a differentiable adaptive fusion network for importance allocation through attention-based TD3 with a hierarchical reward to generate more accurate combination weights.
    \item We evaluate DIMM's tracking performance on various collected 3D trajectory datasets, demonstrating its effectiveness in tracking accuracy improvement and excellent generalization.
    
\end{itemize}


%% file: sec/2_related_work.tex
\section{Related work}
\label{sec:relat_work}

\begin{figure}[h]
  \centering
\includegraphics[width=1.05\linewidth]{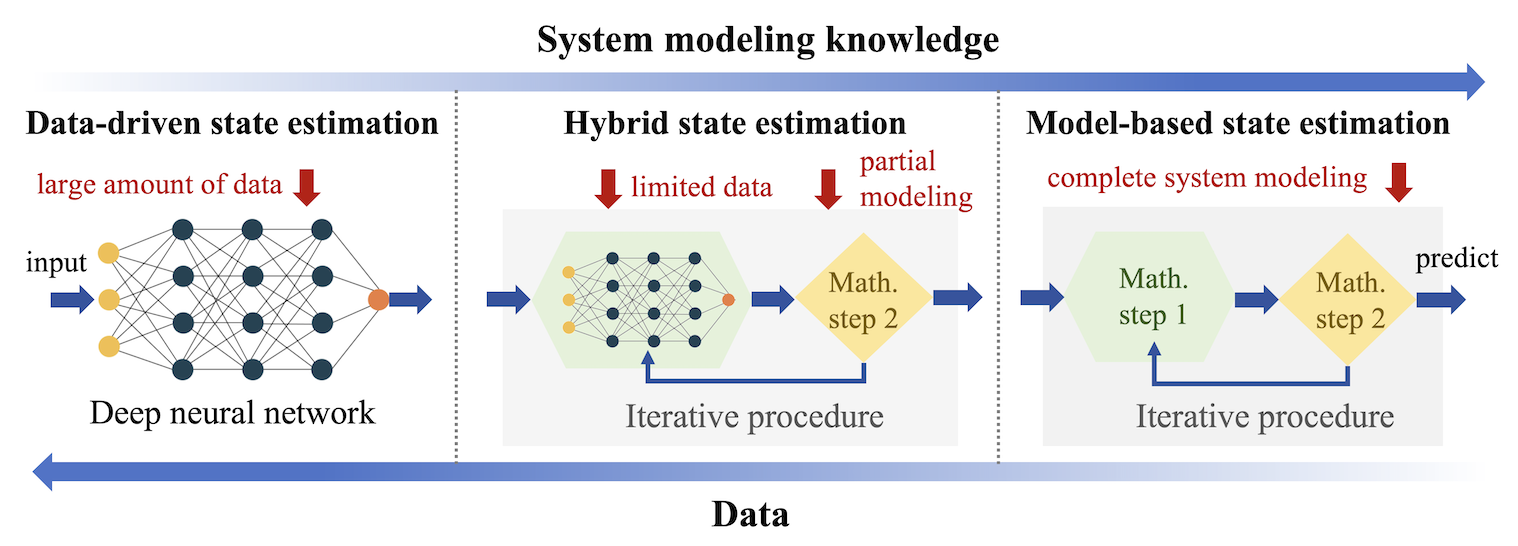}
   \label{fig:related_work}
   \vspace{-0.45cm} 
   \caption{Relationship between three kinds of methods.}
   \vspace{-0.45cm}
\end{figure}

Existing work of state estimation for object tracking lies in three categories, model-based, data-driven, and hybrid ones, as illustrated in \cref{fig:related_work}.

\textbf{Model-based state estimation} for object tracking relies on certain domain knowledge, i.e., prior knowledge of the dynamic system's physical modeling, including the target's movement function and measurement equation. Common model-based estimation methods involve Kalman filter (KF) \citep{kalman1960new} for linear systems with Gaussian noises. To deal with nonlinear problems, more sophisticated filters such as extended Kalman filter (EKF) \cite{ribeiro2004kalman}, unscented Kalman filter (UKF) \cite{liu2007hierarchical}, and cubature Kalman filter (CKF) 
\cite{arasaratnam2009cubature} 
are proposed. For nonlinear systems containing non-Gaussian noises, particle filter (PF) \cite{wang2005online} is further developed based on random sampling. Compared to single-filtering approaches, IMM is designed for target tracking with high maneuverability and unknown dynamics \cite{jilkov1999design}, confronting limitations of the fixed motion model representation and increasing the estimation accuracy by model mixing. 
Overall, model-based methods offer interpretability through explicit physical models \cite{jin2021new}, finding applications in tracking, navigation, and pose estimation \cite{rawlings2012optimization}. However, their performance degrades with inaccurate models in complex systems. Thus, our algorithm enhances model characterization by combining multiple filter estimates.

\textbf{Data-driven state estimation} emerges for object tracking with unknown dynamics as deep learning technology matures, 
requiring no system modeling knowledge compared to model-based techniques. 
Data-driven state estimation is broadly divided into non-parametric and parametric methods. Non-parametric approaches, such as Gaussian Processes (GPs) \cite{seeger2004gaussian}, provide flexible modeling of state and measurement dynamics but often require computational approximations, like sigma or inducing points, for longer sequences. 
Parametric methods primarily utilize deep neural networks (DNN), particularly recurrent architectures like RNNs \cite{medsker2001recurrent} and LSTMs \cite{hochreiter1997long}, which require supervised learning with access to true state information \cite{coskun2017long, shi2021incorporating, jin2024sampled}. Recent advancements, including the dynamical variational autoencoders (DVAEs) \cite{krishnan2017structured} and Kalman variational autoencoder (KVAE) \cite{fraccaro2017disentangled}, enable unsupervised learning by combining a VAE with a linear Gaussian state-space model \cite{girin2020dynamical}. More recent approaches, such as the Recurrent Kalman Network (RKN) \cite{becker2019recurrent} and DANSE \cite{ghosh2024danse}, integrate neural networks with Bayesian techniques, providing a balance between analytical tractability and estimation performance.
Data-driven methods can extract features from measurements even when complex dynamic systems are difficult to model \cite{shlezinger2023model}. However, purely data-driven state estimation requires substantial data and computational resources, while neural network approaches often lack interpretability. Therefore, we adopt network-aided estimation and design learning-based adaptive estimate combination.


\textbf{Hybrid state estimation} is developed for object tracking with partially known dynamics, which integrates both the model-based and data-driven methods. 
KalmanNet \cite{revach2022kalmannet} is a typical hybrid estimation approach using a recurrent neural network (RNN) to model the Kalman gain, which is a supervised learning scheme trained by true states and noisy measurements. For unlabeled training data with only observation values, unsupervised KalmanNet \cite{revach2022unsupervised} is proposed. Split-KalmanNet \cite{choi2023split} is further developed to compensate for the state and measurement model mismatch effects through two parallel networks. Recently, an optimized KF (OKF) \cite{greenberg2024optimization} is developed by optimizing the process and measurement noise covariance matrices, and it is validated that OKF outperforms the Neural KF (NKF) \cite{de2020normalizing} with LSTM sequential model. Existing works also incorporate IMM with DNN. Representative ones include the improved LSTM-based IMM \cite{deng2020improved} and XGBoost-based IMM \cite{chen2016xgboost}, which utilize different models to predict the model interaction weights. 
%
Hybrid state estimation combines model knowledge and data learning, reducing data requirements while improving accuracy \cite{shlezinger2023model}. Therefore, we propose an accurate hybrid state estimation method that balances both the training data amount and system modeling prerequisite well. Particularly, we design a decoupled model-based IMM as our algorithm's framework and adopt an improved reinforcement learning (RL) module to generate the adaptive model combination weights in our work.

%% file: sec/3_problem_formulation.tex
\section{Problem formulation}
\label{sec:pro_form}
In the 3D object tracking problem, the sensor's noisy measurements serve as input, and the estimated state of the target is produced as output. A discrete-time object tracking system \cite{zha2023privacy} with diverse dynamic models is formulated as
\begin{equation}
	\left\{
	\begin{aligned}
		\boldsymbol{x}_k&=f^i\left( \boldsymbol{x}_{k-1}\right) +\boldsymbol{w}_{k-1}^i,  \\
		\boldsymbol{z}_k&=h^i\left( \boldsymbol{x}_k\right) +\boldsymbol{v}_k^i, \ \forall i\in\boldsymbol{\mathcal{M}},
	\end{aligned}
	\right.
	\label{eq track_model}
\end{equation}
where $\boldsymbol{x}_k\in\mathbb{R}^n$ represents the target's state at time step $k$, $\boldsymbol{z}_k\in\mathbb{R}^m$ denotes the measurement, and $\boldsymbol{\mathcal{M}}=\left\{m_1,m_2,...,m_M\right\}$ is the model set. Function $f^i(\cdot): \mathbb{R}^n\rightarrow\mathbb{R}^n$ specifies the target's state transition equation, and $h^i(\cdot): \mathbb{R}^n\rightarrow\mathbb{R}^m$ is the sensor's measurement equation, both of which vary with the model $i\in\boldsymbol{\mathcal{M}}$\footnote{Note that the state dimension $n$ and measurement dimension $m$ also vary with different motion and observation models, respectively. In this paper, all models’ measurements are set as the object’s noisy 3D position as the sensor’s observation transformation is not our main focus.}. Process noise $\boldsymbol{w}^i_{k-1}\in\mathbb{R}^n$ and measurement noise $\boldsymbol{v}_k^i\in\mathbb{R}^m$ are model-dependent with Gaussian distributions following $\boldsymbol{w}^i_{k-1}\sim\mathcal{N}(\boldsymbol{0},\boldsymbol{Q}^i_{k-1})$ and $\boldsymbol{v}_k^i\sim\mathcal{N}(\boldsymbol{0},\boldsymbol{R}_k^i)$, respectively, where $\boldsymbol{Q}^i_{k-1}$ and $\boldsymbol{R}_k^i$ are the corresponding noise covariance matrices. 
Specifically, we denote the Markov transition probability of a model jump process from model $i$ to $j$ as $\pi^{ij}$.
Commonly utilized motion models in IMM include constant velocity (CV) model $m_{cv}$ and constant acceleration (CA) model $m_{ca}$ for linear movements, and constant turn rate (CT) model $m_{ct}$ for nonlinear dynamics, which are detailed in the Appendix.



%% file: sec/4_methodology.tex
\section{Methodology}
\label{sec:method}

\subsection{Revisiting interacting multiple model}
\label{sec:IMM_limits}
As a popular yet effective way to track the target with high maneuverability, IMM algorithm \cite{mazor1998interacting} combines multiple motion models, including CV, CA, and CT models,  simultaneously to estimate the object's state, adapting to different movement patterns by weighting each model's predictions based on their likelihood with respect to measurement. 
Each iteration of the IMM algorithm includes four steps: interaction, filtering, weight generation, and combination, as shown in 
the Appendix\footnote{Only the simplest case using Kalman Filter (KF) is considered in 
the IMM algorithm,
where the linear state transition and measurement function of model $j$ are denoted as $\boldsymbol{F}^j$ and $\boldsymbol{H}^j$, respectively. For details about IMM with more sophisticated filters like Extended Kalman Filter (EKF) and Unscented Kalman Filter (UKF) that are applicable to nonlinear systems, one may refer to \citet{mazor1998interacting}. }.
However, as mentioned above, two critical limitations exist in the combination step of IMM, i.e., the planar solution space constraint (\textbf{L1}), and the observation-dependent weight instability (\textbf{L2}).


\subsubsection{Planar solution space constraint}
\label{sec:IMM_limit1}

\begin{figure}[htbp]
  \centering
  \includegraphics[width=1\linewidth]{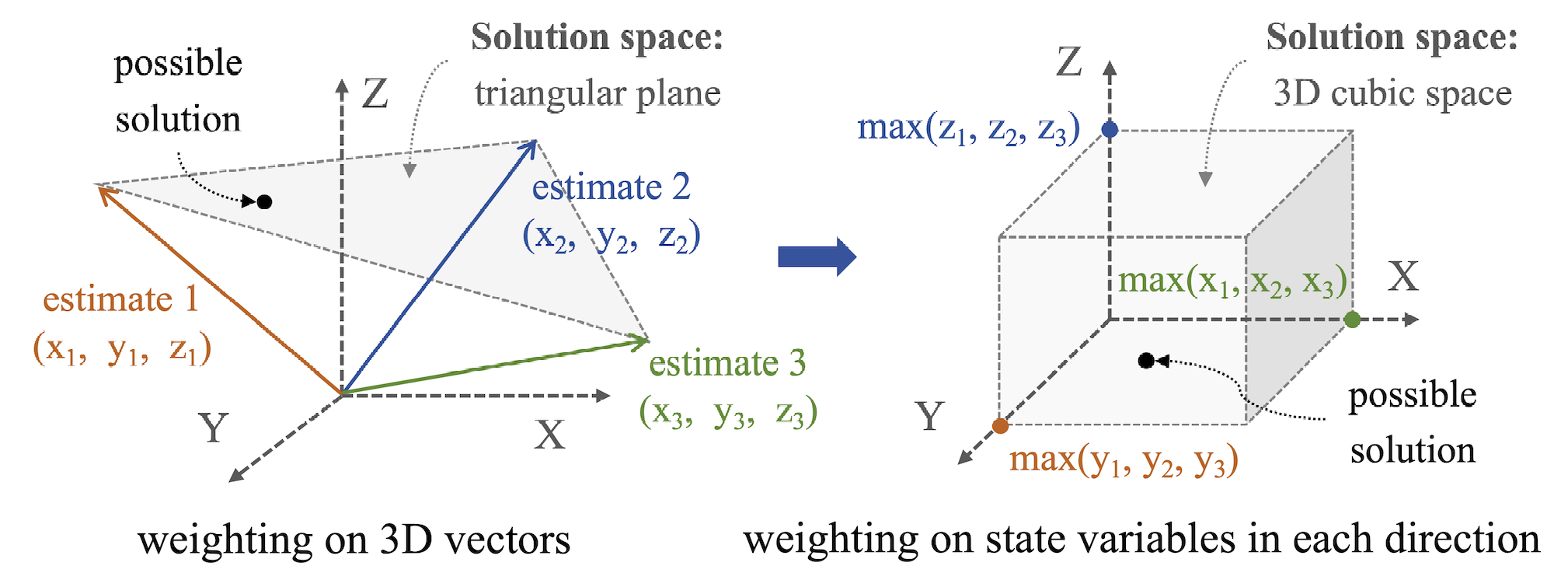}
   \caption{Extend the combination solution space by converting the weighting object from state vectors to variables in each direction.}
   \label{fig:solution_space}
   \vspace{-0.45cm}
\end{figure}

The conventional IMM algorithm \cite{mazor1998interacting} implements direct weighting of the state estimates in all three directions obtained from $M$ different motion models with an $M$-dimensional combination weight vector. Nonetheless, in cases where the target's motion model differs in each direction, such combination operation is no longer optimal. Actually, the multi-model state estimation can be regarded as a 3D convex optimization problem, while the traditional IMM algorithm restricts the feasible domain to a triangular planar region, extremely limiting the optimizable range of the solution space, as shown in \cref{fig:solution_space}.
\begin{proposition}
    \label{pro: solution_space}
    The solution space of IMM's estimate combination is a hyperplane, while the solution space of 3D model combination weights is a hypercube.
\end{proposition}
\begin{proof}
    The proof of \cref{pro: solution_space} is given in 
    the Appendix.
\end{proof}

\noindent\textbf{Solution for L1.} Existing work based on IMM relies on nonlinear models to describe the object's complex movement, which results in interactions between different dimensional variables, thus preventing the independent model recognition and fusion for each direction. Therefore, we aim to design a multi-hierarchy linear filter bank for various motion models that can realize the 3D decoupling of the target's movements to facilitate the independent linear combination of the state's variable in each direction, which is addressed in \cref{sec:solution1}. 
Moreover, to cater to the need for expanded 3D combination solution space, we consider a weight matrix rather than a weight vector for more reasonable estimate combination, as specified in \cref{sec: transformation_matrix}.


\subsubsection{Observation-dependent weight instability}
\label{sec:IMM_limit2}
As a crucial part of IMM, the method of model combination weight generation
significantly affects the estimation accuracy. 
The classic IMM algorithms \cite{mazor1998interacting} compute model combination weights based on the observation likelihood under the Gaussian distribution assumption, which may not be accurate enough as the measurements themselves are subject to errors, especially in cases of frequent measurement loss, high observation noises, and non-Gaussian noise distributions. Moreover, the transition probability function predefined manually in the interaction step of IMM is also uncertain, bringing instability to the model switching at each time step. 
\\
\textbf{Solution for L2.} Since a learnable model recognition approach is needed for more accurate model selection and fusion to align with the target's current movement mode, we propose an adaptive fusion network with TD3 (AdaFuse-TD3) in \cref{sec:solution2} to decide the combination weight matrix rather than relying on the mathematical observation likelihood function for more accurate estimate combination.

\subsection{Overview of our DIMM approach}

\begin{figure*}[htbp]
  \centering
  \includegraphics[width=1\linewidth]{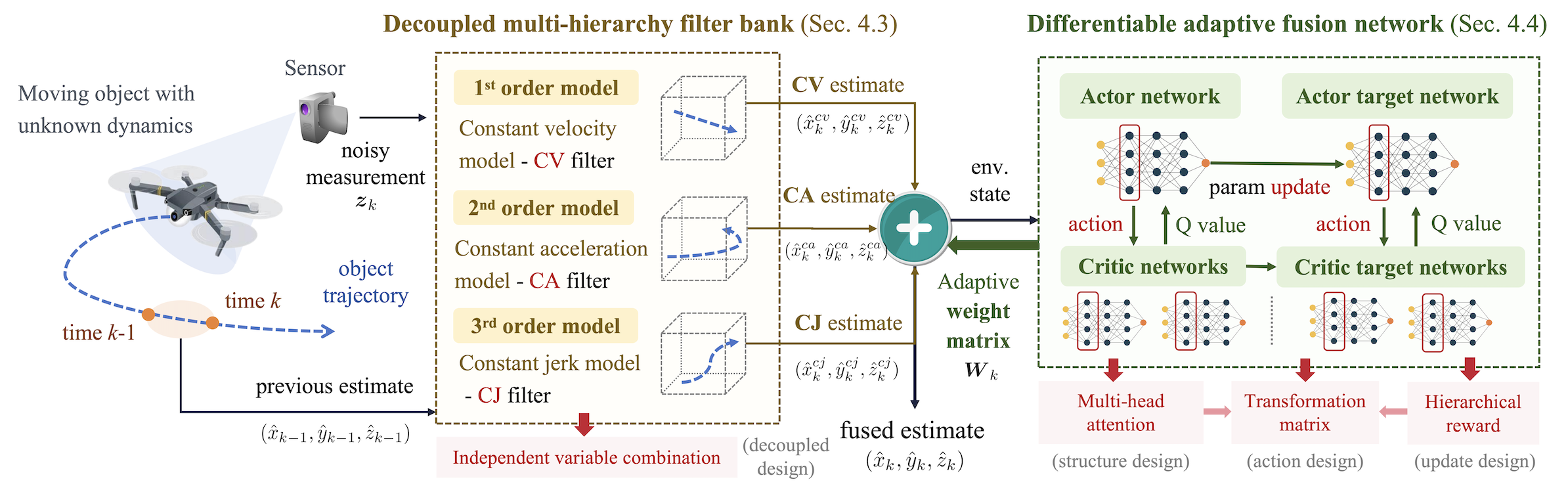}
   \caption{\textbf{Overview of DIMM.} Some critical technical contributions are highlighted in red.
   }
   \label{fig:overview}
   \vspace{-0.45cm}
\end{figure*}

DIMM contains two main modules, a \textit{decoupled multi-hierarchical filter bank} (DHFB) for multi-order local estimation, and a \textit{differentiable adaptive fusion network} (DAFN) for multi-model estimate fusion, as depicted in \cref{fig:overview}. 

\begin{itemize}[leftmargin=*]
    \item \textbf{DHFB module} uses a multi-order motion model group to describe the object's movements, where each model runs a separate KF to generate its local state estimates. Our designed filter bank enables an independent linear combination of the filter's estimate variables in each spatial dimension, thus spanning a larger combination solution space for subsequent estimate fusion. 
    \item \textbf{DAFN module} employs an attention-based TD3 architecture with a hierarchical reward to recognize motion patterns and assign importance weights to each model's estimates for subsequent fusion. Specifically, we take sequential measurements and multi-model estimates as the network input and obtain the transformation matrix of each model as the output to determine the model's combination weights in different dimensions.
\end{itemize}

Finally, the weighted combination of estimates is able to produce the fused object tracking result. The two modules' innovative design is displayed in \cref{fig:overview}. Our proposed DIMM algorithm is specified 
in the Appendix. 

\subsection{Decoupled multi-hierarchical filter bank}
\label{sec:solution1}
Our model is built on a 3D-decoupled multi-hierarchy filter bank with a model group $\boldsymbol{\mathcal{M}}_{D}$, composed of the CV, CA, and constant jerk (CJ) model, to describe the object's movements. By considering various motion models with different orders, our filter bank not only facilitates the independent model combination of dimension-specific motion in each direction, but also provides a more accurate representation of highly nonlinear 3D movements than existing methods based on the CT model\footnote{The conventional model group including CT models relies on strong idealized assumptions about the target's circular motion, such as a fixed turning rate. However, these assumptions are unsuitable for highly nonlinear scenarios, such as emergency stops or abrupt, uneven turns.}, thus improving the estimation accuracy. 

According to the state vector considered, the CV, CA, and CJ model correspond to the first-order, second-order, and third-order motion model, respectively\footnote{The multi-order models' mathematical representation is detailed in the Appendix.}. 
Specifically, our multi-order motion model group is completely made up of linear models, which brings convenience for the subsequent separate linear weighted fusion of the state vector's 3D components, hence realizing the state decoupling in three directions. 
%

%

Therefore, the DHFB module based on the basic KF is effective enough for our model's state estimation, further simplifying the algorithm's computation complexity. Then, one obtains the posterior state estimate $\boldsymbol{\hat x}_k^i$ of each motion model as
\begin{equation}
    \begin{aligned}
        \boldsymbol{\hat x}_k^i=&\boldsymbol{\hat x}_{k|k-1}^i+\boldsymbol{K}_k^i(\boldsymbol{z}_k-\boldsymbol{\hat z}_k^i) \\
    =&f^{i}(\boldsymbol{\hat x}_{k-1}^i)+\boldsymbol{K}_k^i(\boldsymbol{z}_k-h^i(f^{i}(\boldsymbol{\hat x}_{k-1}^i)), \\ 
    i&\in\boldsymbol{\mathcal{M}}_D,
    \end{aligned}
    \label{eq: pos_est}
\end{equation}
where the model group $\boldsymbol{\mathcal{M}}_D=\{m_{cv}, m_{ca}, m_{cj}\}$, $\boldsymbol{\hat x}_{k|k-1}^i$ represents the prior state estimate of model $i$ at time step $k$, $\boldsymbol{\hat z}_k^i$ refers to the predicted measurement, and $\boldsymbol{K}_k^i$ denotes the Kalman gain.

Based on the 3D-decoupled multi-hierarchy filter bank, the problem now turns to recognizing the most appropriate order of the motion models and amplifying its output impact in the combination step to better fit the target's movement pattern for each direction.


\subsection{Differentiable adaptive fusion network }
\label{sec:solution2}
To address the challenges mentioned above, rather than combining models' estimates from weight vectors calculated by the observation likelihood function, we generate the transformation matrix for each model through our designed RL module, AdaFuse-TD3, to meet the needs of motion pattern recognition and adaptive combination weight adjustment at each time step in three directions. 
By learning from interaction with the environment, the fusion network generates transformation matrices that are adaptive to the unpredictable and dynamic behaviors of the object over time. 
Particularly, the transformation matrix is seen as an importance allocation metric for each motion model as it decides the interaction weight value in model combination. 


\subsubsection{Environment definition}
\label{sec: environment_definition}
The position estimation environment of AdaFuse-TD3 can be seen as a Markov decision process (MDP) represented by a tuple $(\mathcal{S}, \mathcal{A}, \mathcal{R}, \mathcal{P}, \gamma)$, where $\mathcal{S}$ is the state space, $\mathcal{A}$ is the action space, $\mathcal{R}$ is the reward, $\mathcal{P}$ is the transition probability distribution, and $\gamma\in [0,1)$ is the discount factor. We define the environment elements as follows.
\\
\textbf{State space} $\mathcal{S}: \boldsymbol{s}_k=[\boldsymbol{z}_{k-l:k}; \boldsymbol{\hat p}_k^{m_{cv}}; \boldsymbol{\hat p}_k^{m_{ca}}; \boldsymbol{\hat p}_k^{m_{cj}};
\boldsymbol{\hat p}_k]\in\mathbb{R}^{15}$. The state of our environment includes the $l$-length measurement sequence\footnote{For the time step $k<l$, 
we perform a zero-padding operation on the missing measurement dimensions.
}, filtered position estimates of the multi-hierarchy filter bank, and the fused position estimate.
\\
\textbf{Action space} $\mathcal{A}: \boldsymbol{a}_k=[\boldsymbol{a}_{k,x}; \boldsymbol{a}_{k,y}; \boldsymbol{a}_{k,z}]\in\mathbb{R}^9$, where $\boldsymbol{a}_{k,j}=[a_{k,j}^{m_{cv}}, a_{k,j}^{m_{ca}}, a_{k,j}^{m_{cj}}]^{\rm T}, j\in\{x,y,z\}$. We take the change of the importance weight value of the decoupled multi-hierarchy filter bank as the action and compute the corresponding transformation matrix of each model based on the action values, as given in \cref{sec: transformation_matrix}. 
\\
\textbf{Reward} $\mathcal{R}: r_k\in\mathbb{R}$. We take the difference of the localization error between our algorithm and a benchmark filter as a hierarchical reward, which is detailed in \cref{sec:reward_design}.
\\
\textbf{Agent.}
We adopt a decision model inspired by TD3 \cite{fujimoto2018addressing} for weight values generation with continuous action space and design an improved network structure as specified in \cref{sec: network_structure}.

\subsubsection{Attention-based network structure}
\label{sec: network_structure}
\begin{figure*}[htbp]
  \centering
  \includegraphics[width=1\linewidth]{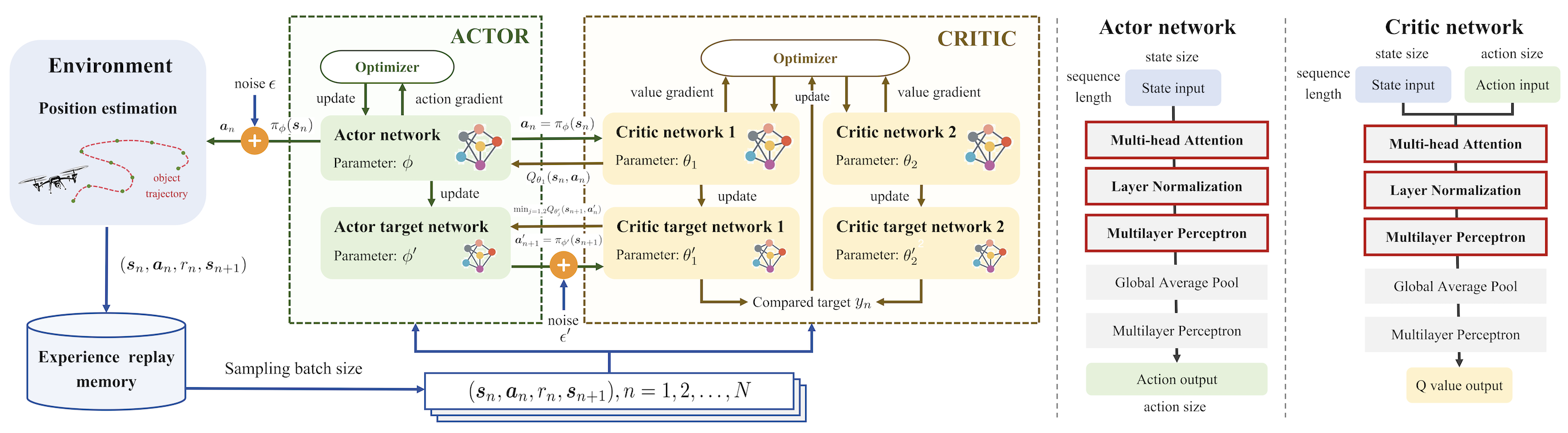}
   \caption{Network structure of the DAFN module.
   }
   \label{fig:network_structure}
   \vspace{-0.45cm}
\end{figure*}

Considering the input measurements are time-sequential and inter-correlated, we build the actor-critic network based on the multi-head attention mechanism \cite{vaswani2017attention} to effectively capture long-range motion patterns. The network architecture is depicted in \cref{fig:network_structure}. Unlike LSTM networks \cite{hochreiter1997long} that process sequences sequentially, our attention-based structure enables parallel processing of temporal dependencies across the entire sequence. Then, the attention-encoded motion features are fed into subsequent multilayer perceptrons to generate the importance weight matrices. 

\subsubsection{Transformation matrix construction}
\label{sec: transformation_matrix}
To facilitate the decoupled combination of filters' position estimates in 3D space, we construct a diagonal transformation matrix for each model as
\begin{equation}
    \boldsymbol{T}^i_k=\text{diag}\{w_{k,x}^i, w_{k,y}^i, w_{k,z}^i\}, i\in\boldsymbol{\mathcal{M}}_D,
    \label{eq: transformation_matrix}
\end{equation}
where the 3D importance weight matrix follows
\begin{equation}
        \boldsymbol{W}_{k}=
        \begin{pmatrix}
            (\boldsymbol{w}_{k,x})^{\rm T}\\
            (\boldsymbol{w}_{k,y})^{\rm T}\\
            (\boldsymbol{w}_{k,z})^{\rm T}
        \end{pmatrix}
        =\begin{pmatrix}
            w_{k,x}^{m_{cv}} & w_{k,x}^{m_{ca}} & w_{k,x}^{m_{cj}} \\
            w_{k,y}^{m_{cv}} & w_{k,y}^{m_{ca}} & w_{k,y}^{m_{cj}}\\
            w_{k,z}^{m_{cv}} & w_{k,z}^{m_{ca}} & w_{k,z}^{m_{cj}}
        \end{pmatrix}.
    \label{eq: importance_matrix}
\end{equation}
Specifically, the weight value of model $i\in\boldsymbol{\mathcal M}_D$ in each direction $j\in\{x,y,z\}$ is generated as a constant between $[0,1]$ through a softmax function according to
\begin{equation}
    w_{k,j}^{i}=\frac{e^{a_{k,j}^i-\parallel \boldsymbol{a}_{k,j}\parallel_{\infty}}}{\sum_{j\in\{x,y,z\}} e^{a_{k,j}^i-\parallel \boldsymbol{a}_{k,j} \parallel_{\infty}}},
    \label{eq: importance_weight}
\end{equation}
where 
$\boldsymbol{a}_k=[\boldsymbol{a}_{k,x}; \boldsymbol{a}_{k,y}; \boldsymbol{a}_{k,z}]\in\mathbb{R}^9$
denotes the action vector as defined in \cref{sec: environment_definition}.
Then, we combine the estimate of each model based on its corresponding transformation matrix and obtain the fused position estimate as
\begin{equation}
    \boldsymbol{\hat p}_k=\sum_{i\in\boldsymbol{\mathcal{M}_D}}\boldsymbol{T}_k^i\boldsymbol{\hat p}_k^i,
    \label{eq: combination}
\end{equation}
where $\boldsymbol{\hat p}_k^i$ is the position variable in estimate $\boldsymbol{\hat x}_k^i$.
Note that only the communal position estimate of the multi-hierarchy filter bank is considered for combination in this paper to simplify the fusion process, as the state dimension varies with motion models\footnote{For more rigorous combination of all state variables of various motion models with unequal state dimension, one can refer to \citet{zubavca2022innovative}. Moreover, we only consider the combination of position estimate in this paper since the object's position information is enough in most practical application scenarios.}.  

\subsubsection{Hierarchical reward design}
\label{sec:reward_design}
Most of the existing RL-aided KF research \cite{gao2020rl, tang2021reinforcement} uses the opposite of the localization error as a reward according to
\begin{equation}
    r_k = -\parallel \boldsymbol{p}_k-\boldsymbol{\hat p}_k \parallel_2,
    \label{eq: sim_reward}
\end{equation}
where $\boldsymbol{p}_k$ is ground truth of the object's position.
%
However, the estimation error in \cref{eq: sim_reward} is highly susceptible to unknown environmental noises, which may cause instability to the convergence of reward. 
Therefore, we intend to reduce the reward variance by designing a hierarchical reward calculated from the difference between the filtering error of AdaFuse-TD3 and that of another benchmark filtering result and feeding it back as an advantageous signal into the network. Specifically, we define the hierarchical reward as:
\begin{equation}
    r_k = -\parallel \boldsymbol{p}_k-\boldsymbol{\hat p}_{k,\text{AdaFuse-TD3}} \parallel_2 + \parallel \boldsymbol{p}_k-\boldsymbol{\hat p}_{k, \text{IMM}} \parallel_2,
    \label{eq: our_reward}
\end{equation}
where $\boldsymbol{\hat p}_{k,\text{AdaFuse-TD3}}$ denotes the position estimation results of our filtering method based on AdaFuse-TD3, and $\boldsymbol{\hat p}_{k,\text{IMM}}$ is the estimate obtained from the non-learning IMM approach.
In this case, the larger the reward value, the higher the estimation accuracy of the filtering method based on AdaFuse-TD3, and the learning-based algorithm outperforms the benchmark when the reward value in \cref{eq: our_reward} is positive.
In summary, the hierarchical reward design weakens the effect of ambient noises, thus flattening the signal wave and improving the model convergence. 

%% file: sec/5_experiments.tex
\section{Experiment}
\label{sec:exp}

\subsection{Experimental setup }

\subsubsection{Datasets}
\textbf{OKF dataset} \cite{greenberg2024optimization} is a driving trajectory dataset consisting of segments with diverse accelerations and turn radius\footnote{
Note that we reconstructed and preprocessed OKF
data using publicly released code in \cite{greenberg2024optimization} since the original dataset is not open.}.
\\
\textbf{Multi-model dataset} is a self-built target trajectory dataset composed of random combinations of trajectory sequences generated by different motion models' dynamics, including CV, CA, CJ, and CT models.
\\
\textbf{Flightmare dataset} is a drone trajectory dataset featuring randomly generated velocities in three directions, collected from Flightmare\footnote{The trajectory collected from Flightmare is regarded as realistic, as it accounts for practical factors such as the drone's dynamic characteristics in the process of motion generation.}, a versatile and high-fidelity quadrotor platform for real-world validation.
\\
\textbf{Lorenz attractor dataset} \cite{revach2022kalmannet, ghosh2024danse, revach2022unsupervised} is a time-series 3D chaos dataset commonly used for algorithms testing in dynamic systems\footnote{The evaluation results for the Lorenz attractor dataset are illustrated in the Appendix due to the limited space.}.

\subsubsection{Baselines and metrics}
\begin{itemize}[leftmargin=*]
    \item \textbf{KF} \cite{kalman1960new} is a classic state estimation method.
    \item \textbf{IMM} \cite{mazor1998interacting} is a famous technique for target tracking with unknown switching dynamic models.
    \item\textbf{RKN} \cite{becker2019recurrent} (Recurrent Kalman Network) is an end-to-end learning approach for KF.
    \item \textbf{DANSE} \cite{ghosh2024danse} (Data-driven Nonlinear State Estimation) is the state-of-the-art model-free method.
    \item \textbf{LSTM-IMM} \cite{deng2020improved} is an IMM method based on LSTM.
    \item \textbf{XGBoost-IMM} \cite{li2021improved} is a XGBoost-based IMM method.
    \item \textbf{OKF} \cite{greenberg2024optimization} is an optimized KF with parameter learning.
    \item \textbf{Mean squared error} (MSE) is an evaluation metric suitable for undesirable large-error cases.
    \item \textbf{Mean absolute error} (MAE) is an indicator of estimation accuracy preferable for required robustness to outliers.
\end{itemize}


\subsection{Estimation accuracy}

\subsubsection{Quantitative results}
\textbf{DIMM outperforms existing state-of-the-art state estimation
methods in terms of estimation accuracy. }
To quantify the algorithm's performance on object tracking accuracy, the MSE and MAE of estimates obtained from baselines and DIMM are compared in \cref{tab: estimation_acc}.
Results show that the performance of model-based algorithms like KF and IMM varies with datasets. 
%
%
Specifically, the UKF-based IMM fails in the OKF dataset due to the numerical sensitivity, suggesting its significant reliance on the operating scenario \cite{seah2011algorithm}.
From \cref{tab: estimation_acc}, we can tell that DIMM is the most accurate tracking scheme compared with existing state-of-the-art state estimation,
which confirms the effectiveness of DIMM in accurate 3D object tracking. 

\renewcommand{\arraystretch}{1}
\begin{table*}
  \centering
  \small
    \caption{\textbf{Comparison of estimation errors of DIMM with seven baselines.} The results are averaged over 100 randomized trials. 
    }
  \scalebox{0.85}
  {
  \begin{tabular}{cccccccccc}
    \toprule
    \multirow{2}{*}{Datasets} &  \multirow{2}{*}{Metrics} & \multicolumn{2}{c}{Model-based methods} & \multicolumn{2}{c}{Data-driven methods} & \multicolumn{4}{c}{Hybrid methods} \\
    \cline{3-10}
    & & KF \cite{kalman1960new} & IMM \cite{mazor1998interacting} & RKN \cite{becker2019recurrent} & DANSE \cite{ghosh2024danse} & LSTM-IMM \cite{deng2020improved} & XGBoost-IMM \cite{li2021improved} & OKF \cite{greenberg2024optimization} & DIMM (ours)
    \\ \midrule
     \multirow{2}{*}{OKF data} 
     & MSE & 5.1771 & -  & 0.6132 & 0.6408 & 0.9053 & 3.5254 & 3.9890 & \textbf{0.4431} \\
     & MAE & 3.1713 & - & 0.1835 & 0.1687 & 0.2052 & 2.6929 & 1.6090 & \textbf{0.1124}
    \\ \midrule
    \multirow{2}{*}{Multi-model data} 
     & MSE & 2.7535 & 2.0290 & 0.7442 & 0.0310 & 1.8879 & 3.4526 & 3.9509 & \textbf{0.0041} \\
     & MAE & 2.1202 & 1.7635 & 0.1373 & 0.1430 & 4.6926 & 2.2531 & 1.5643 & \textbf{0.0542}
    \\ \midrule
    \multirow{2}{*}{Flightmare data} 
     & MSE & 129.0360 & 129.5573 & 1.7353 & 1.6920 & 2.9830 & 5.5448 & 7.8423 & \textbf{1.4934} \\
     & MAE & 101.5394 & 102.2903 & 1.1978 & 1.2630 & 4.0271 & 3.7056 & 3.2317 & \textbf{1.0100}
     \\
    \bottomrule
  \end{tabular}
   }
  \label{tab: estimation_acc}
  \vspace{-0.45cm}
\end{table*}

\subsubsection{Qualitative results}
\begin{figure}[H]
  \centering
  \begin{subfigure}{0.29\linewidth}
    \includegraphics[width=1.05\linewidth]{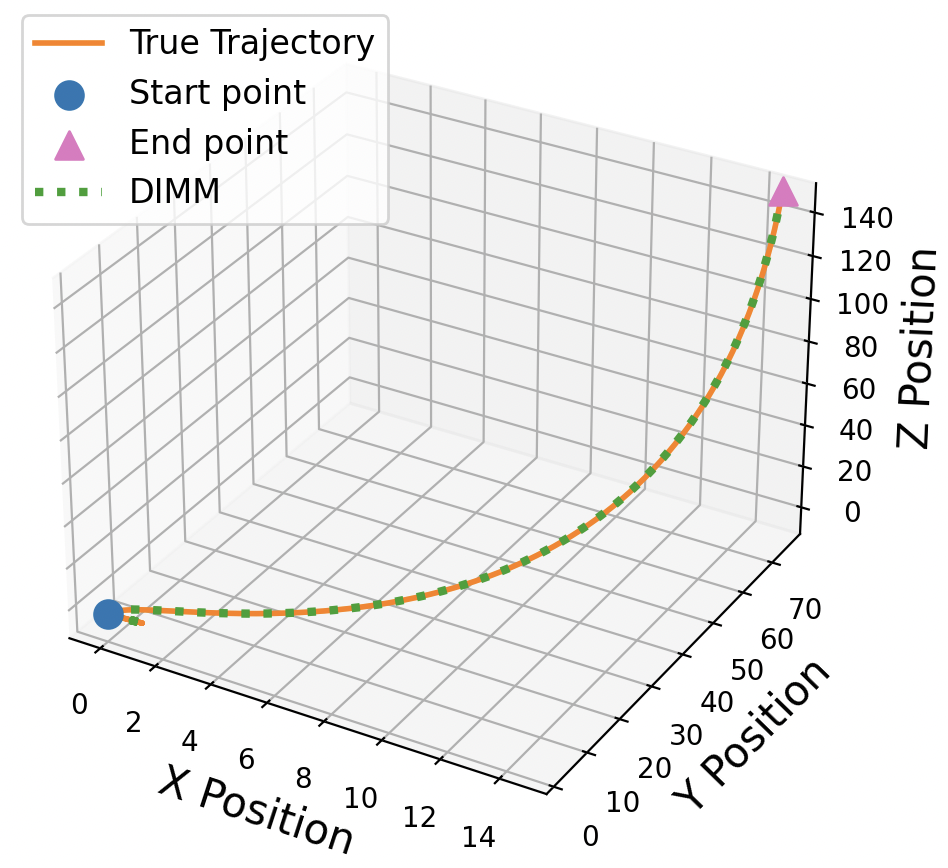}
    \caption{Multi-model data.}
    \label{fig:track_mydata}
  \end{subfigure}
  \hfill
  \begin{subfigure}{0.28\linewidth}
  \includegraphics[width=1.06\linewidth]{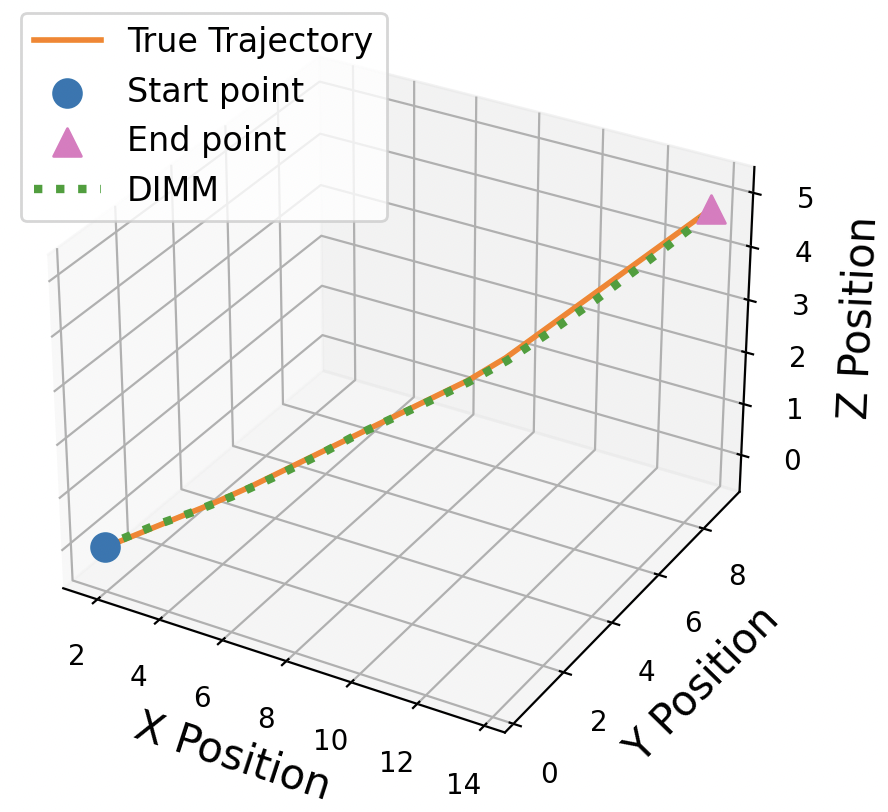}
    \caption{OKF data.}
    \label{fig:track_OKFdata}
  \end{subfigure}
  \hfill
  \begin{subfigure}{0.28\linewidth}
  \includegraphics[width=1.1\linewidth]{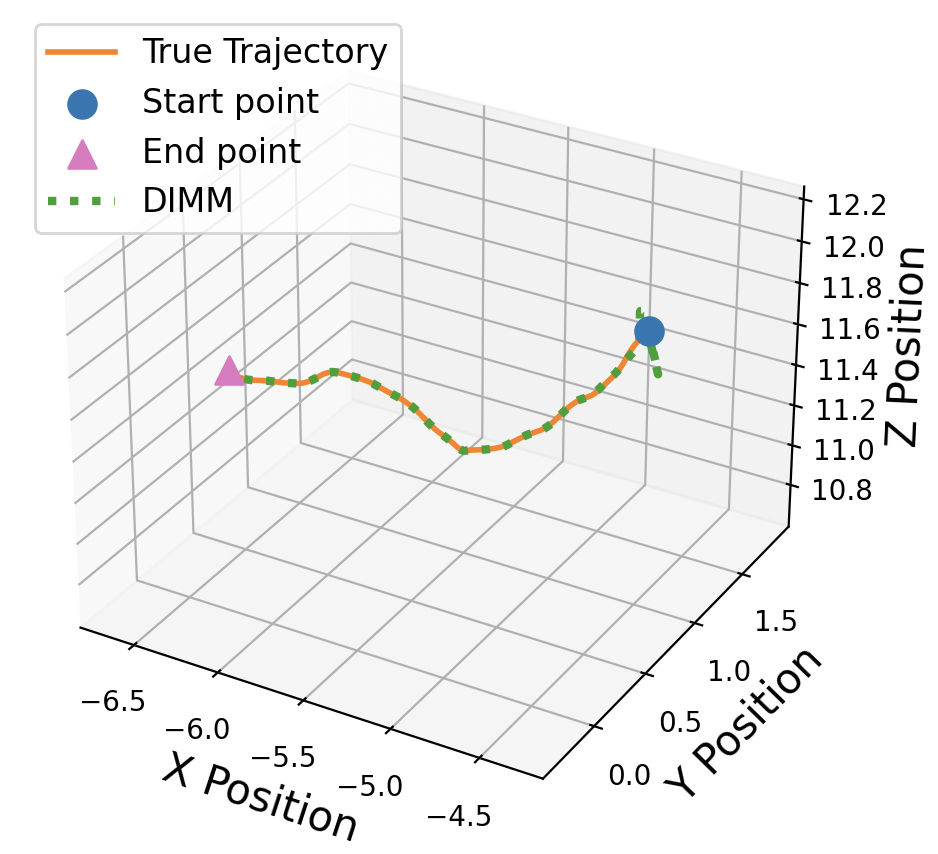}
    \caption{Flightmare data.}
    \label{fig:track_Flightmaredata}
  \end{subfigure}
  
  \caption{Examples of comparison between the actual and estimated trajectories of DIMM.}
  \label{fig:track}
  \vspace{-0.25cm}
\end{figure}

\begin{figure}[H]
  \centering
  \includegraphics[width=1\linewidth]{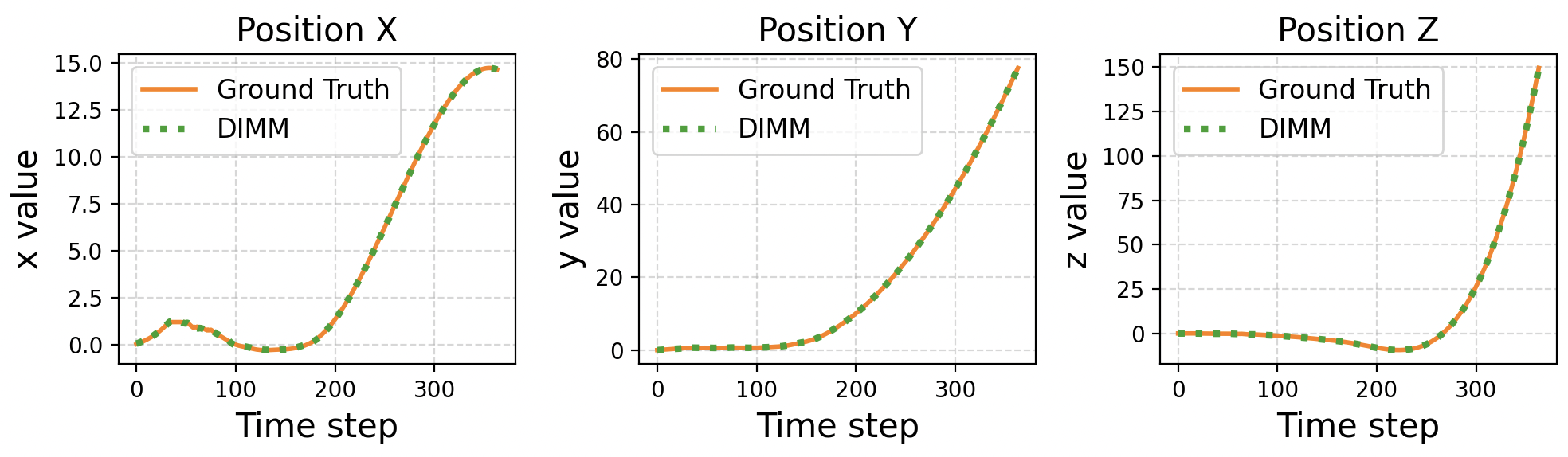}
   \caption{Example of comparison between the actual and estimated position state variables of DIMM.}
   \label{fig:state_pos}
\end{figure}

\textbf{The estimate results of DIMM approximate the true values well.} We compare the object’s ground-truth and estimated trajectory of DIMM in \cref{fig:track}.
It can be seen that our algorithm effectively fits the complex 3D motion trajectory of the target with unknown dynamics.
To further validate the feasibility of our algorithm’s
position estimates, \cref{fig:state_pos} compares the true object position states with the estimated ones obtained from DIMM. As shown, the estimated position variables converge to the ground-truth values, indicating DIMM is applicable to nonlinear 3D object tracking.
  
%
%

\subsection{Study of hierarchical reward} 
\textbf{Hierarchical reward design improves the estimation accuracy.} 
To validate the effectiveness of the reward design illustrated in \cref{sec:reward_design}, we compare the tracking accuracy of our algorithm with and without the hierarchical term in \cref{tab:reward_design}. Specifically, we 
refer the reward defined by \cref{eq: sim_reward} as the simple reward, and our hierarchical reward is given in \cref{eq: our_reward}. It can be seen from \cref{tab:reward_design} that the hierarchical reward design effectively improves the estimation accuracy.
%
%
\renewcommand{\arraystretch}{1}
\begin{table}[t!]
  \centering
  \caption{Estimation errors of DIMM with different rewards.}
  \scalebox{0.68}
  {
  \begin{tabular}{cccccc}
    \toprule
    Reward design & Metrics & OKF data & Multi-model data & Flightmare data \\
    \midrule
    \multirow{2}{*}{DIMM (simple)} & MSE & 0.5389 & 0.1656 & 7.9213
    \\ 
    & MAE & 0.5281 & 0.3507 & 2.5474  
    \\
    \midrule
    \multirow{2}{*}{DIMM (hierarchical)} & MSE & \textbf{0.4431} & \textbf{0.0041} & \textbf{1.4934}  
    \\ 
    & MAE & \textbf{0.1124} & \textbf{0.0542} & \textbf{1.0100} 
    \\
    \bottomrule
  \end{tabular}
  }
  \label{tab:reward_design}
\end{table}
%
\vspace{-0.35cm}

\subsection{Interpretable analysis}
\begin{figure}[htbp]
  \centering

   \begin{subfigure}{1\linewidth}
    \includegraphics[width=1\linewidth]{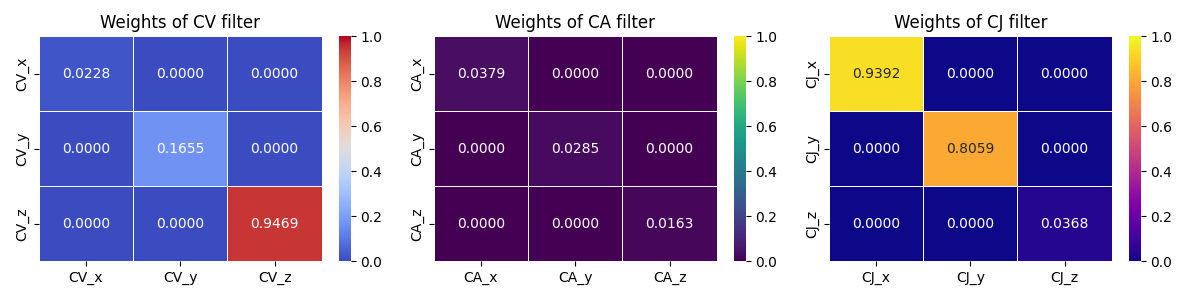}
    \caption{Transformation matrix of each model at one time step on the OKF dataset. It can be seen from the maximum diagonal elements of the transformation matrices that the best-fit models for X, Y, and Z directions are CJ, CJ, and CV model, respectively.}
    \label{fig:transformation_matrix_OKFdata}
  \end{subfigure}
  
  \begin{subfigure}{1\linewidth}
    \includegraphics[width=1\linewidth]{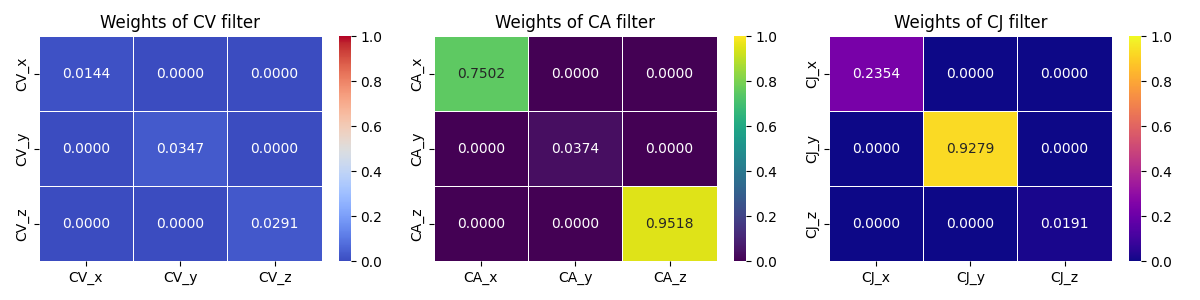}
    \caption{Transformation matrices on the Flightmare dataset. The best-fit models for X, Y, and Z directions are CA, CJ, and CA model, respectively.}
    \label{fig:transformation_matrix_flightmaredata}
  \end{subfigure}

  \begin{subfigure}{1\linewidth}
    \includegraphics[width=1\linewidth]{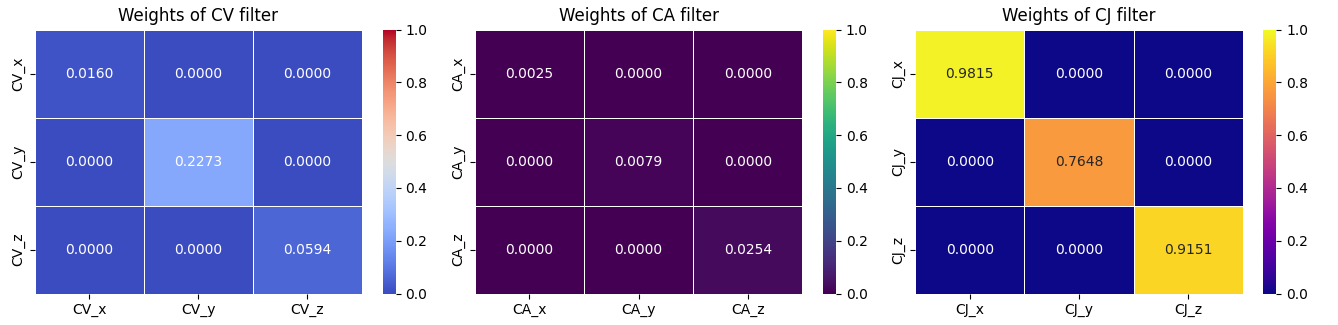}
    \caption{Transformation matrices on our dataset. The best-fit models for X, Y, and Z directions are all CJ model, demonstrating our algorithm's applicability to isotropic single motion pattern.}
    \label{fig:transformation_matrix_ourdata}
  \end{subfigure}
  
  \caption{Examples of the transformation matrix of each motion model's filter from the decoupled multi-hierarchy filter bank.}
  
  \label{fig:interpret_analysis}
  \vspace{-0.3cm}
\end{figure}

\textbf{Transformation matrix is for model importance allocation during combination.} 
For more intuitive understanding of the transformation matrix $\boldsymbol{T}_k^i$ demonstrated in \cref{sec: transformation_matrix}, we depict this diagonal matrix of each model $i\in\boldsymbol{\mathcal M}_D$ at given time steps in \cref{fig:interpret_analysis}.
As seen, the diagonal elements of each model's transition matrix correspond to the fusion weights of each filter's estimate, i.e. the values of $(w_{k,x}^i,w_{k,y}^i,w_{k,z}^i)$ in \cref{eq: transformation_matrix}. According to \cref{eq: combination}, the greater the weight value of the filter with its corresponding model in one direction, the more significant its estimate is during model combination in that direction. Therefore, one can deduce the most appropriate model type of the moving object for each direction of the 3D space in the designed multi-hierarchy filter bank from the transformation matrix of each model at each time step, as analyzed in \cref{fig:interpret_analysis}\footnote{The key point is that we do not rigidly assume that the object’s motion at a given moment is solely characterized by a single motion model in a specific direction, as analyzed in \cref{sec:IMM_limit1}. Instead, we adopt a predicted weighted combination of multiple motion models to enhance the algorithm’s ability to describe certain unknown complex motions.}.

\subsection{Inference efficiency}
\textbf{DIMM demonstrates impressive inference efficiency.} 
When operating on one A800 GPU, DIMM processes batches of 256 in just 22 ms with a 2000 MiB memory footprint. Therefore, DIMM's ability to handle large batches quickly suits real-time and high-throughput tasks, crucial for rapid decision-making in areas like autonomous systems. Overall, our model's outstanding inference efficiency highlights its potential to significantly enhance the performance and efficiency of various applications that demand both speed and accuracy. In conclusion, DIMM's speed and low memory usage enable real-time use and scalability, giving it a competitive edge.

\subsection{Ablation study}
This section conducts an ablation study on the action space size and the DAFN module in \cref{sec:solution2} to evaluate their impacts on our algorithm.
The action space size determines the granularity of possible actions available to the agent, which may influence the accuracy of weight values in our problem. 
Moreover, DAFN module is essential in DIMM as it decides the combination weights for each model's estimate variable in each direction.

\noindent \textbf{The action space size affects the algorithm's estimation accuracy.} The action space size corresponds the range of combination weight values of transformation matrices in our case. We evaluate the effects of different sizes of action space on the position estimation accuracy of DIMM, as shown in \cref{tab: action_space}. It turns out that a larger action space provides finer weight values but increases training complexity, while a smaller action space may limit the model's capability to learn nuanced behaviors.

\renewcommand{\arraystretch}{1}
\begin{table}[t!]
  \centering
  \caption{Estimation errors of DIMM with different action space. 
  }
  \scalebox{0.73}
  {
  \begin{tabular}{ccccccccc}
    \toprule
    \multirow{2}{*}{Action space} & \multicolumn{2}{c}{OKF data} & \multicolumn{2}{c}{Multi-model data} & \multicolumn{2}{c}{Flightmare data}   \\
    \cline{2-7}
      & MSE & MAE & MSE & MAE & MSE & MAE  \\
    \midrule
    $(-5,5)$ & 0.4622 & 0.1563 & \textbf{0.0041} & \textbf{0.0542} & \textbf{1.4934} & \textbf{1.0100} & 
    \\ \midrule
    $(-4,4)$ & 0.4512 & 0.1353 & 0.0065 & 0.0735 & 1.6404 & 1.0154 & 
    \\ \midrule
    $(-3,3)$ & 0.4478 & 0.1276 & 0.0188 & 0.1059 & 1.6134 & 1.0131 & 
    \\ \midrule
    $(-2,2)$ & \textbf{0.4431} & \textbf{0.1124} & 0.0082 & 0.0684 & 1.5890 & 1.0153 & 
    \\ \midrule
    $(-1,1)$ & 0.4519 & 0.1483 & 0.0229 & 0.1121 & 1.6400 & 1.0153 & 
    \\  
    \bottomrule
  \end{tabular}
  }
  \label{tab: action_space}
\end{table}




\noindent \textbf{DAFN module significantly improves the algorithm's estimation accuracy.} From \cref{tab: ablation_RL}, it can be seen that the incorporation of the DAFN module effectively improves the model's performance across all datasets, with distinct reductions in both MSE and MAE metrics. Particularly, 
there exists a respective $88.33\%$, $99.79\%$, and $98.84\%$ reduction in the MSE of DIMM compared to the one without DAFN,
demonstrating the crucial role of the DAFN module in enhancing estimation accuracy. This confirms the advantages of learning-based fusion weight generation over mathematical formula-based generation.

\renewcommand{\arraystretch}{1}
\begin{table}[t!]
  \centering
  \caption{Estimation errors of DIMM w/ and w/o DAFN.}
  \scalebox{0.68}
  {
  \begin{tabular}{cccccc}
    \toprule
    Module setting & Metrics & OKF data & Multi-model data & Flightmare data  \\
    \midrule
    \multirow{2}{*}{DIMM (w/o DAFN)} 
    & MSE & 3.7969 & 1.9824 & 129.0346 
    \\ 
    & MAE & 2.3349 & 1.7045 & 101.5391 
    \\
    \midrule
    \multirow{2}{*}{DIMM (w/ DAFN)} 
    & MSE & \textbf{0.4431} & \textbf{0.0041}  & \textbf{1.4934}  
    \\ 
    & MAE & \textbf{0.1124}  & \textbf{0.0542} & \textbf{1.0100}  
    \\
    \bottomrule
  \end{tabular}
  }
  \label{tab: ablation_RL}
  \vspace{-0.45cm}
\end{table}

%% file: sec/6_conclusion.tex
\section{Conclusion and future work}

This paper proposes a novel 3D object tracking framework, DIMM, for accurate object tracking with unknown dynamics. DIMM consists of a decoupled multi-hierarchy filter bank for multi-order local estimation, which expands the model combination solution space and a differentiable adaptive fusion network, which produces more accurate weights for model combination. Evaluation results on multiple datasets show that our solution significantly improves the object tracking accuracy compared with the SOTA approaches.
As for future work, we plan to deploy the algorithm's applications to real-world systems.


